%% file: main.tex
\documentclass[10pt,twocolumn]{article}

\usepackage[margin=0.72in]{geometry}
\usepackage{booktabs}
\usepackage{amsmath,amssymb}
\usepackage{graphicx}
\usepackage{microtype}
\usepackage[hidelinks]{hyperref}
\usepackage{xcolor}
\usepackage{enumitem}
\setlist{nosep,leftmargin=*}

\title{Do Geometry-Aware Positional Encodings Help Transformers\\
in Spatial Imperfect-Information Games?}
\author{Wenji Fu\\
Research Institute of Economics and Management\\
Southwestern University of Finance and Economics\\
\texttt{fuwenji61616@gmail.com}}
\date{August 2026}

\begin{document}
\maketitle

\begin{abstract}
Transformers applied to spatial imperfect-information games must represent map
geometry while tracking hidden entities through time. We ask whether
geometry-aware positional encodings improve these capabilities, without
claiming a new positional encoding. We construct a four-level benchmark on a
hexagonal naval pursuit game: controlled geometry and topology probes, an
exact-Bayes hidden-target tracking task, offline policy imitation at 1k and
10k games, and 7,200 fixed-seed games against three legacy opponents. Across
matched Transformer backbones, HexRoPE reduces exact-belief posterior
cross-entropy relative to no positional encoding by 0.278 on D6-transformed
test orbits and 0.329 on a larger map; both hierarchical-bootstrap confidence
intervals exclude zero, and both Holm-adjusted $p$-values are below 0.001. At 1k games, HexRoPE improves
policy action accuracy by 4.63 percentage points over no encoding and 2.05
points over rectangular relative bias; the gains shrink to 1.55 and 0.41
points at 10k games. However, HexRoPE does not improve aggregate gameplay win
rate: its paired effect over no encoding is $-1.56$ points (95\% CI
$[-4.50,1.17]$). Rectangular relative bias is strongest on D6 belief
consistency but fails sharply when extrapolating from radius 3 to radius 4,
while graph bias provides only a small blocked-edge gain. The results show
that geometric inductive bias improves belief estimation and data-efficient
imitation, but those representation gains do not automatically produce
stronger closed-loop play.
\end{abstract}

\section{Introduction}

Spatial imperfect-information games combine two problems that Transformers do
not solve merely by flattening a board into a token sequence. First, legal
movement, search radius, and interception depend on direction, distance, and
map topology. Second, the position of an opponent must be inferred from a
history of positive and negative observations. A positional encoding may help
the attention mechanism represent the former, yet it is unclear whether that
advantage survives temporal belief tracking or changes actual play.

Two-dimensional rotary encodings and graph-relative biases already provide
plausible mechanisms for injecting spatial structure
\cite{su2024roformer,he2024ropevit,ostmeier2024liere,ying2021graphormer}. Their evaluation,
however, is usually centered on vision or graph prediction. A game offers a
useful stress test because it separates several notions often conflated as
``spatial understanding'': recovering geometric relations, maintaining a
calibrated posterior over a hidden target, imitating spatial actions, and
winning under closed-loop distribution shift.

We study these notions in a reproducible hex-map naval pursuit environment.
Our aim is empirical rather than architectural novelty: we do not claim to
invent HexRoPE. We compare matched Transformer backbones with no positional
encoding, rectangular relative bias, axial 2D RoPE, cube-coordinate HexRoPE,
and HexRoPE with graph-distance bias. The core of the benchmark is an
exact-belief task whose labels are computed by rule enumeration rather than by
legacy AI actions. It therefore tests hidden-target tracking independently of
policy imitation quality.

Our contributions are:
\begin{itemize}
  \item a four-level benchmark that separates hex geometry, blocked-edge
  topology, exact hidden-state belief tracking, policy imitation, and gameplay;
  \item an exact-Bayes dataset with IID, D6-transformed, larger-map, and
  blocked-edge test sets, including transform and probability-integrity tests;
  \item controlled 1k/10k policy experiments and 7,200 fixed-seed games with
  per-game outputs and hierarchical paired-bootstrap inference; and
  \item an empirical boundary: geometry-aware encodings improve beliefs and
  imitation, especially under low data, but do not reliably improve win rate.
\end{itemize}

\section{Related Work}

\paragraph{Spatial positional encoding.}
Relative position biases inject displacement-dependent terms into attention
logits and are widely used in vision Transformers
\cite{wu2021rethinking,liu2021swin}. RoPE-ViT adapts rotary embeddings to
images, while LieRE casts rotary encoding through group representations
\cite{he2024ropevit,ostmeier2024liere}. More recent multidirectional and
hexagonal variants further motivate testing coordinates beyond a rectangular
axis decomposition \cite{liu2026spiral,byeon2026hexst}. We treat these as
existing mechanisms and study their behavior in an imperfect-information
decision setting.

\paragraph{Graph structure and factorized attention.}
Graphormer and GRPE use shortest-path or relation information as attention
biases \cite{ying2021graphormer,park2022grpe}; comparative studies show that
the usefulness of graph positional encodings depends on the task
\cite{black2024comparing,grotschla2024benchmarking}. This motivates separating
continuous hex geometry from blocked-edge topology. Our policy and belief
models factor spatial and temporal attention, following the broad efficiency
pattern of video Transformers \cite{bertasius2021timesformer,arnab2021vivit}.

\section{Benchmark}

\subsection{Game Setting}

The environment is a research-oriented digital adaptation of Michael Smith's
board wargame \emph{Sink the Bismarck!}, originally published by World Wide
Wargames (3W) in 1992. Its principal board and rules reference is the Japanese
edition published by Kokusai-Tsushin in \emph{Command Magazine Japan} no.~19
\cite{smith1992sinkbismarck}. The software formalizes and modifies parts of the
game for simulation and machine-learning experiments; it is an independent,
unofficial research implementation.

The game is a two-player naval pursuit game on an irregular hex map.
The German side attempts to escape or score objectives; the British side
searches for and intercepts hidden German ships. A policy observes only
side-visible information. Its tensor history has shape $[T,C,H,W]$, and legal
actions include map-targeted movement and search as well as special actions.
Play lasts at most 18 turns. The German ships move while hidden (Bismarck moves
up to two sea edges when undamaged and one when damaged), attack convoys on
sea-route cells for stochastic victory points (VP), and may reveal their
location through radio interception. British ships move up to three reachable
sea edges; co-location reveals German ships, while the carrier Ark Royal may
search one adjacent cell once per turn. A revealed co-location can trigger
combat, whose dice resolve damage and sinking. Germany wins immediately at 6
VP, or by reaching Brest while ahead on VP; Britain wins by sinking Bismarck
or preventing a German victory through turn 18. These asymmetric movement,
search, and victory rules make the hidden location strategically relevant.
The game is used as a controlled scientific instrument, not as a claim of
broad game-playing generality.

\subsection{Geometry and Topology Probes}

The canonical probe enumerates all 1,521 query--target pairs on the 39-cell
irregular game map and predicts direction, hex distance, and blocked-edge BFS
distance. A second probe uses all 1,369 pairs on a regular radius-3 hexagon;
splits are assigned by D6 orbit so transformed copies cannot cross train/test
boundaries. Seven encodings are trained for five seeds with a shared 236k
parameter backbone. These probes diagnose direct access to spatial relations,
but do not by themselves establish hidden-state reasoning.

\subsection{Exact-Belief Benchmark}

We generate hidden-target trajectories under a known legal random-walk
transition, including a stay action. At step $t$, the observer receives the
last sighting, elapsed time, legal search region, and either a successful or
failed noisy search. The label is the exact posterior, not a sampled target or
legacy-policy action. For transition matrix $P$ and observation likelihood
$L(o_t\mid x)$, enumeration computes
\begin{align}
\bar b_t(x) &= \sum_{x'} P(x\mid x')b_{t-1}(x'),\\
b_t(x) &= \frac{L(o_t\mid x)\bar b_t(x)}
{\sum_y L(o_t\mid y)\bar b_t(y)}.
\end{align}
Every posterior is checked to sum to one.

This benchmark isolates the pursuit mechanic rather than replaying every game
rule. Its hidden target starts from an observed cell, takes one uniformly
sampled legal neighboring-or-stay transition per step, and is detected with
probability 0.7 when it lies in the searched center-plus-neighbors region. A
success collapses the posterior to the observed target cell; a failure
downweights every searched cell by likelihood 0.3 before normalization. Convoy
combat, VP, and multiple ships are intentionally excluded, so performance can
be attributed to spatial transition and negative-evidence tracking.

Training contains 10,000 radius-3 episodes, with 1,000 development episodes.
Test sets contain 2,000 IID radius-3 episodes, 500 independent radius-3 base
episodes expanded into 6,000 full D6-orbit examples, 2,000 radius-4 episodes,
and 2,000 blocked-edge episodes. Base episode identifiers are disjoint across
splits. This protocol tests transform consistency on unseen trajectories, not
generalization to a subset of group elements withheld during training. D6 transformations act jointly
on maps, observations, hidden positions, and posteriors; applying the inverse
must recover the original tensors within floating-point tolerance.

We report posterior cross-entropy (CE), KL divergence, multiclass Brier score,
top-$k$ posterior mass, expected hex-distance error, 15-bin expected
calibration error, and D6 consistency. For a transform $g$, consistency is
measured by
\begin{equation}
\operatorname{JS}\!\left(f(x),g^{-1}f(gx)\right),
\end{equation}
averaged over transformed copies of the same episode.

\subsection{Policy Imitation and Gameplay}

The clean policy dataset was regenerated after an observer-causality fix. The
10k set contains 10,000 games and 2,112,926 action records; the low-data set
contains 1,000 games and approximately 211k records. Labels are executed
actions from a reproducible pool of state-machine, evolved, scripted, and
random policies, not expert or human demonstrations. Splits are isolated by
game, preventing trajectory leakage.

For gameplay, each of four 10k policy modes and all three training seeds plays
100 games as each side against random, V11 state-machine, and Yanfu scripted
opponents. All policies share initial game seeds within a matchup, yielding
$4\times3\times2\times3\times100=7{,}200$ games. We record win rate, VP
difference, turns, detection turn, search coverage and repetition, search
dispersion, British crowding, German route diversity, direct-F7 movement, and
stay rate.

\section{Models and Protocol}

\paragraph{Encodings.}
\textsc{None} provides no spatial position. \textsc{RectRel} combines learned
row/column absolute embeddings with an attention-logit bias indexed by
rectangular row/column displacement.
\textsc{AxialRoPE} rotates disjoint query/key subspaces using two coordinate
axes. \textsc{HexRoPE} instead uses the three dependent cube coordinates
$(q,r,s)$ with $q+r+s=0$, distributing rotations across those projections.
\textsc{HexRoPE+Graph} additionally learns a bias indexed by graph distance
and edge availability. Invalid cells are masked in every model.

\paragraph{Belief model.}
The exact-belief tracker uses eight observation frames, hidden dimension 96,
three factorized spatial/temporal blocks, four heads, FFN dimension 192, and a
61-cell posterior head. All five encodings use the same training budget:
15 epochs, batch size 64, AdamW, and three seeds. Graph features are dynamic
inputs; target posterior and true hidden position never enter the model input.

\paragraph{Policy model.}
Grid v4 uses history length 8, four factorized blocks, six heads, dimension
192, FFN dimension 512, dropout 0.1, invalid-cell masking, and legal-action
context. The model has approximately 2.02M parameters. All modes use three
epochs, batch size 128, AdamW at $10^{-4}$, 2,000 warmup steps followed by
cosine decay, mixed precision, value-loss weight 0.25, win-auxiliary weight
0.15, and entropy weight 0.005.

\paragraph{Inference.}
Policy metrics are exported per validation game using the identical game-level
split for each training seed. We report action accuracy and CE, value MAE, and
Brier score of the sigmoid-transformed shared value/win scalar. For principal
comparisons, a hierarchical paired bootstrap first resamples training seeds
and then paired games or episodes. We use 5,000 replicates and report 95\%
percentile intervals. Holm correction is applied to the preregistered
HexRoPE--None and HexRoPE--RectRel primary families. Graph-bias comparisons and
gameplay are explicitly exploratory.

\section{Results}

\subsection{Controlled Spatial Probes}

On the canonical irregular map, five-seed HexRoPE obtains 0.984 direction and
0.977 hex-distance accuracy, while graph bias is strongest on blocked-edge BFS
distance (0.927 accuracy and 0.074 MAE). On the regular D6-orbit split,
HexRoPE reaches 1.000 direction and 0.996 hex-distance accuracy. The no-position
model remains near chance on distance in both probes. Thus cube-coordinate
rotation exposes regular hex geometry, whereas explicit graph relations better
represent removed edges.

\input{generated/paper_tables.tex}

\subsection{Exact Beliefs}

\begin{figure*}[t]
\centering
\includegraphics[width=0.86\textwidth]{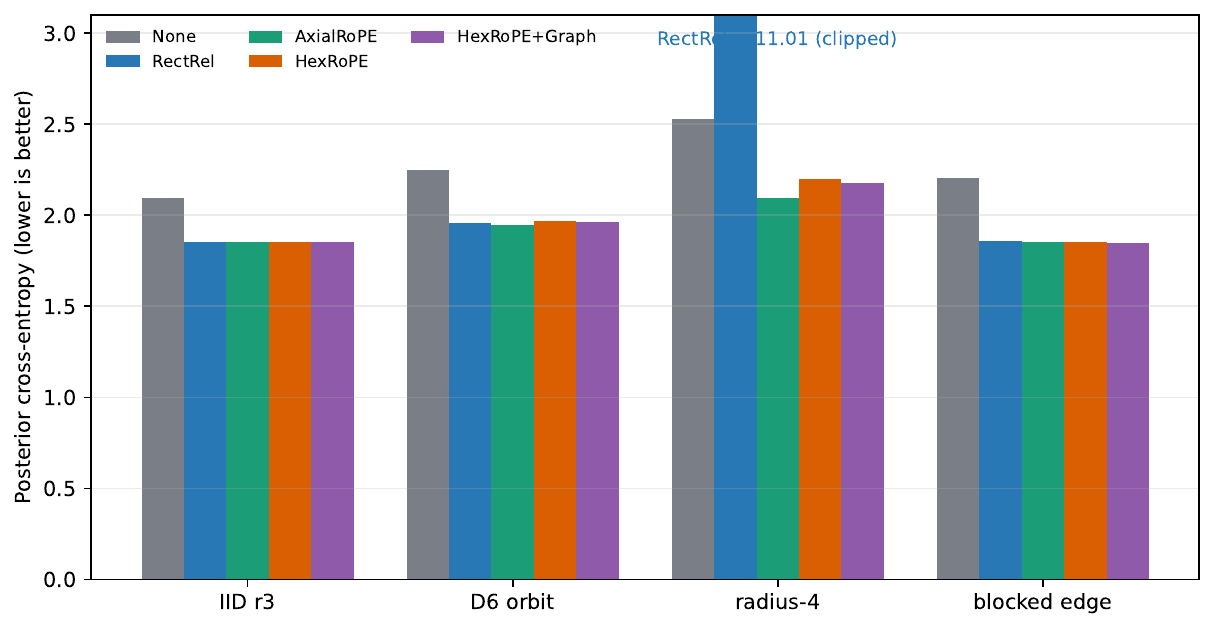}
\caption{Exact-belief posterior CE. RectRel's radius-4 value is clipped for
readability and annotated with its true value.}
\label{fig:belief}
\end{figure*}

Figure~\ref{fig:belief} and Table~\ref{tab:belief} show that all spatial
encodings outperform no encoding. Relative to None, HexRoPE reduces CE by
0.278 on D6 and 0.329 on radius-4; hierarchical 95\% CIs are
$[-0.300,-0.251]$ and $[-0.361,-0.283]$, respectively, with Holm-adjusted
$p<0.001$. This supports the claim that position-aware attention improves
hidden-target posterior estimation beyond imitation labels.

The comparison among spatial encodings is more nuanced. RectRel has the lowest
D6 consistency error (0.0047 versus HexRoPE's 0.0095) and slightly lower D6 CE;
HexRoPE minus RectRel is 0.013 CE, CI $[0.001,0.035]$. Yet RectRel catastrophically
extrapolates to radius 4 (CE 11.008): its radius-4 outer coordinates were never
updated during radius-3 training. Coordinate-based RoPE variants
remain between 2.096 and 2.200. AxialRoPE is best among the tested RoPE forms on
this particular larger-map split. Consequently, our evidence supports
coordinate-based extrapolation, but not a universal ranking of HexRoPE above
all 2D encodings.

Graph bias yields a small blocked-edge improvement over HexRoPE: CE changes by
$-0.00147$, CI $[-0.00202,-0.00083]$, and expected-distance error by
$-0.00349$, CI $[-0.00536,-0.00189]$. Top-1 and Brier intervals include zero.
Topology therefore appears complementary, but the effect is modest in this
benchmark.

\begin{figure*}[t]
\centering
\includegraphics[width=0.98\textwidth]{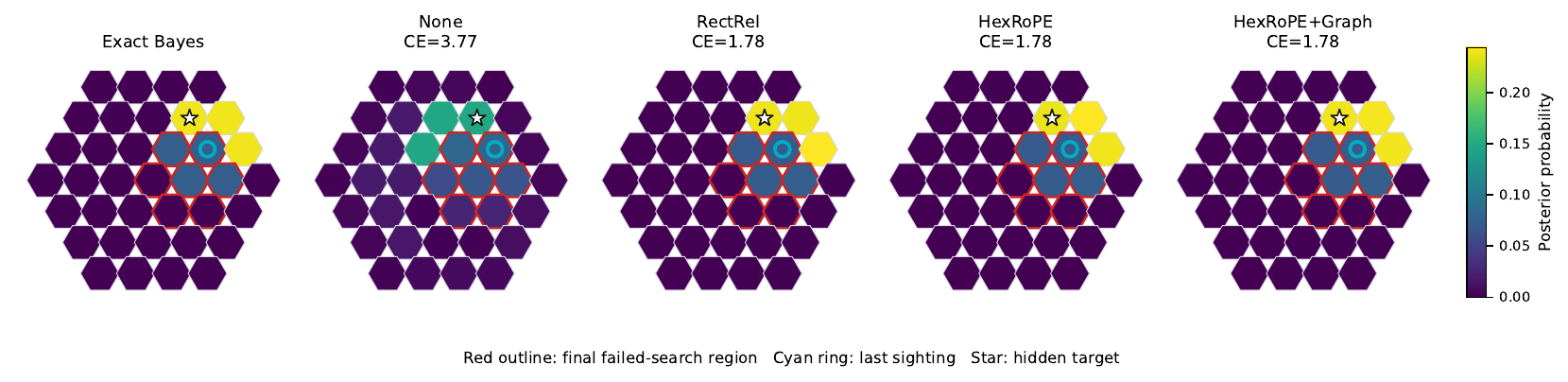}
\caption{Exact-Belief case study after a failed search. Predictions average
three training seeds and share one color scale. Red outlines mark the final
searched region, the cyan ring marks the preceding sighting, and the star
marks the hidden target used only for evaluation. The post hoc selection
criterion required a non-degenerate posterior, a final failed search, and
HexRoPE CE at least 0.7 below None.}
\label{fig:belief-case}
\end{figure*}

Figure~\ref{fig:belief-case} makes the distributional difference concrete. In
episode \texttt{iid\_test\_001256}, a sighting is followed by a failed search.
The exact update leaves seven reachable cells, with three leading modes of
probability 0.238. The seed-averaged HexRoPE variants and RectRel recover this
shape almost exactly, whereas None assigns substantial mass to cells on the
wrong side of the last sighting. This is a post hoc illustrative example with
its selection criterion disclosed, not additional statistical evidence.

\subsection{Data-Efficient Policy Imitation}

\begin{figure}[t]
\centering
\includegraphics[width=\columnwidth]{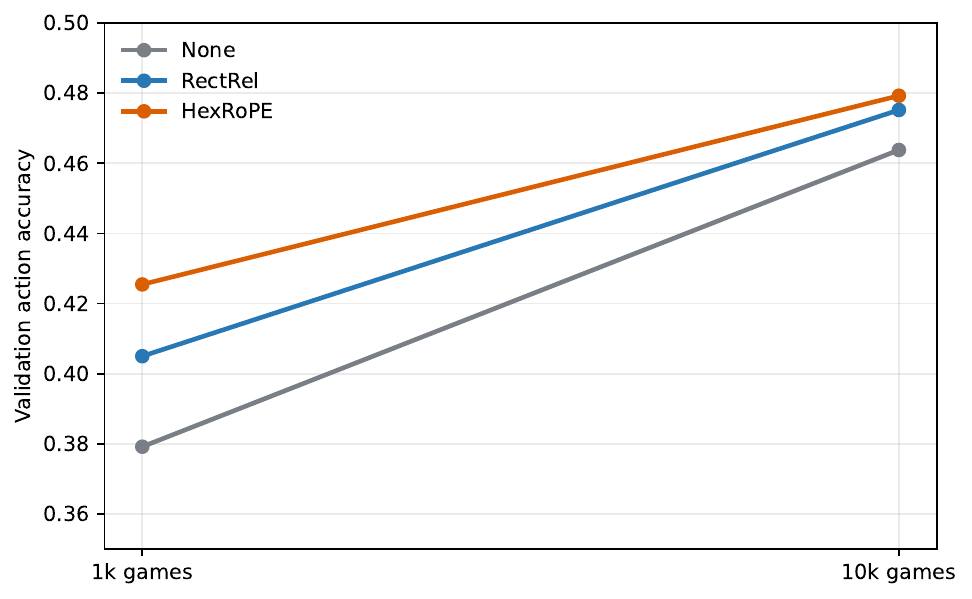}
\caption{Action imitation accuracy at 1k and 10k games. Each point averages
three training seeds; the two points are not presented as a scaling law.}
\label{fig:policy}
\end{figure}

At 1k games, HexRoPE improves action accuracy over None by 0.0463 (95\% CI
$[0.0353,0.0579]$) and over RectRel by 0.0205
($[0.0123,0.0292]$). At 10k games, the effects remain positive but shrink to
0.0155 ($[0.0135,0.0177]$) and 0.00407 ($[0.00289,0.00535]$). All four primary
accuracy tests remain significant after Holm correction. Figure~\ref{fig:policy}
therefore supports a low-data benefit rather than a full scaling-law claim.

Policy CE follows the same direction. Value MAE is not consistently improved
at 1k, and HexRoPE's 1k Brier score is slightly worse than None by 0.00313
($[0.00157,0.00465]$). At 10k, its Brier score is lower by 0.00091. The shared
value/win head should thus not be used as evidence for a robust calibration
advantage.

\subsection{Closed-Loop Gameplay}

Offline improvements do not translate into a reliable aggregate win-rate
gain. Across matched game seeds, HexRoPE minus None is $-0.0156$ in win rate,
with 95\% CI $[-0.0450,0.0117]$. HexRoPE minus RectRel is $-0.0228$
($[-0.0550,0.00722]$), and HexRoPE+Graph minus HexRoPE is $-0.0050$
($[-0.0311,0.0206]$). All intervals cross zero.

The aggregate result hides matchup dependence. HexRoPE is stronger than None
as Germany against V11 (0.727 versus 0.670), but weaker as Britain against
Yanfu (0.337 versus 0.457). It reduces search repetition relative to RectRel by
0.051 ($[-0.0978,-0.0157]$), yet does not significantly increase search
dispersion and slightly increases British crowding relative to None. These
behaviors are consistent with the observed failure mode of concentrating
search assets near recent information. Positional representation alone does
not supply strategic diversity, opponent adaptation, or long-horizon credit.

\begin{figure*}[t]
\centering
\includegraphics[width=0.82\textwidth]{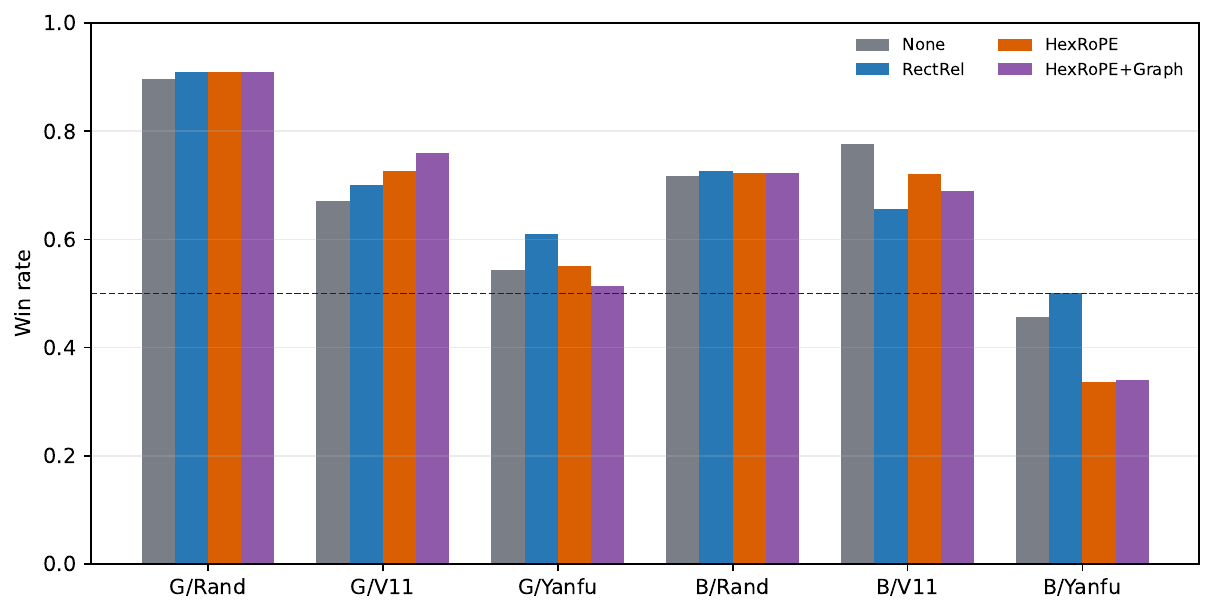}
\caption{Win rates over 300 games per architecture--side--opponent cell.
Dashed line marks 50\%. G/B denote the side controlled by the learned policy.}
\label{fig:gameplay}
\end{figure*}

\section{Discussion}

The four evidence levels answer different questions. Geometry probes establish
that the attention mechanism can recover direction and distance. Exact-belief
tracking shows that this access improves posterior prediction when the labels
come from known dynamics rather than an old AI. Policy imitation shows a
larger benefit under lower data. Gameplay then supplies the missing boundary:
the resulting policy is not reliably stronger in closed loop.

The most defensible interpretation is therefore not ``HexRoPE solves spatial
games.'' It is that explicit geometric structure reduces the representational
and sample burden of spatial prediction, while topology needs a separate bias
and strategy needs additional learning signals. The sharp radius-4 failure of
the learned absolute components in RectRel also cautions against evaluating
only on fixed-size boards. Conversely,
RectRel's strong D6 consistency cautions against treating cube-coordinate RoPE
as automatically equivariant to physical rotations and reflections.

For future game-cognition architectures, the present models provide a
controlled Transformer baseline. A promising next step is to expose exact or
learned beliefs to an action-consequence module and train expected-utility or
counterfactual objectives. That change targets the gap observed here between
representing where the hidden target may be and choosing a strategically
useful action under that uncertainty.

\section{Limitations}

The benchmark uses one game and small hex maps, so external validity remains
limited. Exact-belief dynamics are intentionally controlled and simpler than
the full game. Policy labels come from legacy AIs rather than humans, experts,
or a solved policy; imitation accuracy measures fidelity to that pool. Fixed
seeds align initial randomness, but model actions alter later states, so paired
gameplay does not imply identical state trajectories. Only three training
seeds are available for the expensive policy and gameplay studies, making the
training-seed hierarchy more informative than a naive game-level sample size.
Finally, HexRoPE is an evaluated prior method family, not a novelty claim, and
the recent 2026 preprints should be updated if their archival metadata changes.

\section{Reproducibility}

The exact-belief aggregate SHA-256 begins with
\texttt{b3927015cab6d7df}; the complete value is recorded in
\path{deeplearn/data/paper1_exact_belief/dataset_manifest.json}.
Each split manifest stores episode and map identifiers, transform, generation
seed, legal mask, posterior, and file hash. Every evaluator emits per-game or
per-episode JSON. Formal commands, checkpoints, hardware metadata, and live
state are stored beside results. The main entry points are
\texttt{generate\_belief\_benchmark.py},
\texttt{run\_paper1\_belief\_matrix.py},
\texttt{run\_paper1\_policy\_matrix.py},
\texttt{evaluate\_policy\_checkpoints.py},
\texttt{run\_paper1\_gameplay\_matrix.py}, and
\texttt{build\_paper1\_artifacts.py}. All runners skip completed tasks and
enforce a 5.5-hour launch budget.

\section{Conclusion}

Geometry-aware positional encodings help Transformers estimate exact beliefs
and imitate spatial policies, with the largest policy gain under low data.
Their benefits are conditional: the tested rectangular absolute-plus-relative
encoding is strong in-distribution but fragile across map size, graph bias adds only a small topology gain, and better
representations do not reliably improve win rate. This separation between
representation, belief, imitation, and play is the central empirical result
and a foundation for testing explicit game-cognition modules.

\section*{Acknowledgments}

The author gratefully acknowledges game designer Michael Smith, whose board
wargame \emph{Sink the Bismarck!} provides the foundation for the pursuit
setting and core game mechanics studied here, and Kokusai-Tsushin Co., Ltd.,
publisher of the Japanese \emph{Command Magazine} edition used as the principal
board and rules reference. This research implementation is independent and
unofficial; all software modifications, experimental abstractions, and errors
are the author's own.

\bibliographystyle{plain}
\bibliography{references}
\end{document}

%% file: generated/paper_tables.tex
\begin{table*}[t]
\centering
\caption{Exact-belief tracking (three-seed means). Lower is better for CE, KL, and D6 JS; higher is better for top-1.}
\label{tab:belief}
\small
\begin{tabular}{lrrrrrrrr}
\toprule
& \multicolumn{2}{c}{IID radius-3} & \multicolumn{3}{c}{D6 transforms} & \multicolumn{2}{c}{radius-4 OOD} & blocked \\
\cmidrule(lr){2-3}\cmidrule(lr){4-6}\cmidrule(lr){7-8}\cmidrule(lr){9-9}
Encoding & CE$\downarrow$ & KL$\downarrow$ & CE$\downarrow$ & Top-1$\uparrow$ & JS$\downarrow$ & CE$\downarrow$ & KL$\downarrow$ & CE$\downarrow$ \\
\midrule
None & 2.095 & 0.245 & 2.247 & 0.362 & 0.0172 & 2.529 & 0.471 & 2.203 \\
RectRel & 1.856 & 0.005 & 1.956 & 0.381 & 0.0047 & 11.008 & 8.949 & 1.858 \\
AxialRoPE & 1.853 & 0.003 & 1.949 & 0.378 & 0.0033 & 2.096 & 0.038 & 1.855 \\
\textbf{HexRoPE} & 1.853 & 0.003 & 1.969 & 0.380 & 0.0095 & 2.200 & 0.141 & 1.852 \\
HexRoPE+Graph & 1.853 & 0.002 & 1.962 & 0.372 & 0.0074 & 2.177 & 0.118 & 1.850 \\
\bottomrule
\end{tabular}
\end{table*}

\begin{table}[t]
\centering
\caption{Offline policy imitation at two data scales. Means are over three training seeds.}
\label{tab:policy}
\small
\begin{tabular}{lrrrr}
\toprule
& \multicolumn{2}{c}{1k games} & \multicolumn{2}{c}{10k games} \\
\cmidrule(lr){2-3}\cmidrule(lr){4-5}
Encoding & Acc.$\uparrow$ & CE$\downarrow$ & Acc.$\uparrow$ & CE$\downarrow$ \\
\midrule
None & 0.379 & 1.888 & 0.464 & 1.574 \\
RectRel & 0.405 & 1.910 & 0.475 & 1.541 \\
\textbf{HexRoPE} & 0.425 & 1.794 & 0.479 & 1.521 \\
\bottomrule
\end{tabular}
\end{table}

\begin{table*}[t]
\centering
\caption{Fixed-seed gameplay win rates. Each cell pools 300 games from three training seeds; gameplay is exploratory.}
\label{tab:gameplay}
\small
\begin{tabular}{lrrrrrr}
\toprule
& \multicolumn{3}{c}{German policy} & \multicolumn{3}{c}{British policy} \\
\cmidrule(lr){2-4}\cmidrule(lr){5-7}
Encoding & Random & V11 & Yanfu & Random & V11 & Yanfu \\
\midrule
None & 0.897 & 0.670 & 0.543 & 0.717 & 0.777 & 0.457 \\
RectRel & 0.910 & 0.700 & 0.610 & 0.727 & 0.657 & 0.500 \\
HexRoPE & 0.910 & 0.727 & 0.550 & 0.723 & 0.720 & 0.337 \\
HexRoPE+Graph & 0.910 & 0.760 & 0.513 & 0.723 & 0.690 & 0.340 \\
\bottomrule
\end{tabular}
\end{table*}